\documentclass[journal]{IEEEtran}
\usepackage{graphicx}
\usepackage{ragged2e}
\usepackage{multirow}
\usepackage{amsmath,amsfonts}
\usepackage{algorithmic}
\usepackage{array}
\usepackage{threeparttable}
\usepackage{multirow}
\usepackage[caption=false,font=normalsize,labelfont=sf,textfont=sf]{subfig}
\usepackage{textcomp}
\usepackage{stfloats}
\usepackage{url}
\usepackage{verbatim}
\usepackage{graphicx}
\usepackage[style=base]{caption}
\ifCLASSINFOpdf

\else

\fi

\usepackage[justification=centering]{caption}

\begin{document}
%
\title{Multi-modal Fusion Technology based on Vehicle Information: A Survey}
%
%
%

\author{Yan Gong*, Jianli Lu, Jiayi Wu, Wenzhuo Liu


\thanks{e-mail: gongyan2020@foxmail.com, lujianli364@163.com, wujiayi96@outlook.com, liuwenzhuo3318@gmail.com;}}

%
%

\markboth{}
{Shell \MakeLowercase{\textit{et al.}}: Bare Demo of IEEEtran.cls for IEEE Journals}
%



\maketitle

\begin{abstract}

Multi-modal fusion is a basic task of autonomous driving system perception, which has attracted many scholars' interest in recent years. The current multi-modal fusion methods mainly focus on camera data and LiDAR data, but pay little attention to the kinematic information provided by the bottom sensors of the vehicle, such as acceleration, vehicle speed, angle of rotation. These information are not affected by complex external scenes, so it is more robust and reliable. In this paper, we introduce the existing application fields of vehicle bottom information and the research progress of related methods, as well as the multi-modal fusion methods based on bottom information. We also introduced the relevant information of the vehicle bottom information data set in detail to facilitate the research as soon as possible. In addition, new future ideas of multi-modal fusion technology for autonomous driving tasks are proposed to promote the further utilization of vehicle bottom information.
\end{abstract}

\begin{IEEEkeywords}
Multi-modal fusion, perception, autonomous driving.
\end{IEEEkeywords}

%
\IEEEpeerreviewmaketitle

\section{Introduction}
%
%
%
%

\IEEEPARstart In the early research of semantic segmentation, object detection and other fields, most of them are based on a single sensor such as camera and LiDAR \cite{garcia2017review,zhao2019object}. The camera has the characteristics of a high frame rate and high resolution, which provide accurate information in normal environments, such as good weather and lighting. However, in bright light or dark conditions, such as severe camera exposure or thunderstorm weather, the camera does not provide much information as in the normal environments \cite{bijelic2020seeing}.
LiDAR has a strong ability to acquire three-dimensional information and provide accurate distance measurement, which is not affected by lighting conditions, but greatly affected by weather, e.g., rainy day and foggy day \cite{bijelic2020seeing}. Each sensor has its advantages and disadvantages, so they are inevitably limited by sensors. 

Due to the limitations of a single sensor, multi-sensor fusion is extensively studied to reduce the effects of the hardware itself and the weather.
In recent years, multi-modal fusion methods for autonomous driving perception tasks have developed rapidly. 
There are many types of sensors used in autonomous driving, including monocular vision, stereo vision, LiDAR, vehicle dynamic information obtained from the car's odometer and inertial measurement unit (IMU), and global positioning information obtained using positioning systems and digital map.
In the prediction process, single-modal information only reflect the value of information from a single point of view, while multi-modal information combines information from two or more modalities for prediction, reflecting the value of different information and making the information complementary, make up for the shortcomings of a single sensor \cite{bramon2011multimodal, gao2017discriminative}. 
Therefore, recent works use multi-modal fusion technology to fuse the data of different modalities to play their respective advantages and improve the accuracy and robustness of the prediction results.


At present, in the field of autonomous driving, most methods conduct multi-modal fusion research on point cloud data captured by LiDAR and image data captured by camera, such as fusion camera-lidar for 3D detection and 3D trajectory generation \cite{chen2017multi,asvadi2018multimodal,oh2017object,schlosser2016fusing,wang2017fusing}. The research on the use of vehicle information is still lacking. However, the LiDARs and cameras are sensors external to the vehicle in most tasks, which are affected by factors such as sensor cost, sensor hardware limitations, and weather. Some information from the vehicle itself, e.g., vehicle speed, acceleration, turning angle, and engine current, are accurately obtained in real-time from the vehicle itself and smart portable devices \cite{sathyanarayana2012leveraging}, which are no need to install additional sensors on the vehicle to obtain these vehicle information, greatly reducing sensor cost. However, vehicle information, which is not affected by ambient weather and can be obtained in real time, is not widely used.

Because vehicle information have not been fully utilized, we have introduced all aspects of vehicle information in this article. First, we mainly analyze the existing researches on vehicle information such as steering angle and vehicle speed. The application of different information requires good fusion technology, so we describe some fusion technologies. Then some data set containing vehicle information are introduced to provide sufficient data sources for research. Afterwards, we describe some challenges in applying vehicle information. Finally, we introduce the future research directions in detail, and prove that vehicle information has an important auxiliary role for existing task research through experiments. For example, the steering angle can be applied to steering control or lane recognition during the turning process of autonomous vehicles, which not only avoids the influence of strong light, but also gives the task a prior data. In the process of object tracking, the change degree of the object position can be reflected by the vehicle speed. The faster the vehicle speed is, the greater the change degree of the object is. Therefore, the vehicle speed has a certain auxiliary effect on object tracking.
The speed can also be used to judge the driver's driving style and estimate the expected mileage of the vehicle.
We believe that the vehicle information contained in the underlying vehicle sensors, such as vehicle speed, steering angle, brake and accelerator, are very important for many tasks of automatic driving. Based on this information, extensive research can be carried out in the above areas. In view of this, we have conducted an extensive investigation on the research direction of multi-modal fusion based on vehicle information, and put forward the existing challenges and future research prospects. Ours contributions can be summarized as follows:
\begin{itemize}
    \item [$\bullet$] As far as we know, our paper is the first investigation focusing on the multi-modal fusion of vehicle information, including steering angle prediction, multiple auxiliary tasks, lane change prediction, driving behavior classification, vehicle speed prediction, trajectory prediction, etc.
    \item [$\bullet$] We also introduced and tested two future research directions of vehicle information fusion to prove the role of underlying information, such as: lane detection combined with steering angle and images, and driving behavior classification combined with vehicle speed and images.
    \item [$\bullet$] Our paper organizes different datasets related to vehicle information, and proposes some problems in the research field of vehicle information fusion. In addition, we introduce the fusion method of vehicle information and other information for different data space, which provides a great convenience for other scholars to quickly start the field
   
\end{itemize}

The remaining organization of this paper is as follows: section \ref{2} reviews the research status based on vehicle information. We introduce some multi-modal fusion technologies in Section \ref{3}. Section \ref{4} presents the public datasets containing vehicle information in the field of autonomous driving. Then, we describe the existing challenges and research prospects in Section \ref{5} and section \ref{6} respectively. Finally, the conclusion is drawn in Section \ref{7}.

\section{Represent Methods}
\label{2}
The following mainly introduces the application status and representative methods of steering angle and vehicle speed. In terms of steering angle, we recommend some research on steering angle prediction, multiple auxiliary tasks, lane change prediction and perception under noise image data. In terms of speed, we present some research on driving style classification, speed prediction, trajectory prediction, mileage, time, energy consumption forecasts, ideslip angle, vehicle deceleration prediction.

\subsection{Steering angle}
\subsubsection{Steering Angle prediction}
In recent years, Convolutional Neural Networks (CNNs) have achieved remarkable performance in various perception and control tasks. The key factor behind these impressive results is their ability to learn millions of parameters using large amounts of labeled data. This section focuses on introducing the use of deep learning techniques to combine the underlying information of the vehicle with other sensor information to assist autonomous driving.

First, Bojarski et al. \cite{bojarski2016end} proposed the use of direct mapping of raw pixels from a single front-facing camera to steering commands, an end-to-end approach that proved to be very powerful. Using sensor data from the vehicle's underlying layers, the system learned to drive on roads and highways without lane markings. For areas with insufficient visual information, such as parking lots and unpaved roads. The system automatically learns internal representations for necessary processing steps, such as detecting useful road features, using only the person's steering angle as a training signal. It does not need to be explicitly trained to detect, for example, the contours of roads. Fusing camera data with steering data optimizes all processing steps end-to-end compared to explicit decomposition of problems such as lane marking detection, path planning, and control. Besides, Rausch et al. \cite{rausch2017learning} proposed an end-to-end controller for autonomous vehicles based on convolutional neural network (CNN). The deployed framework does not require explicit human-designed algorithms for lane detection, object detection, or path planning. Comparing the performance of the controller with the steering behavior of a human driver yields good results. 

In addition to vehicle steering issues in normal driving, vehicles may be ambiguous when approaching intersections, or use additional command information to navigate the vehicle, but not efficiently. Aiming at the problem of vehicles traveling automatically under a given path, Wang et al. \cite{wang2019end} proposed a new effective navigation command(sub-target angle), which does not require human participation and is calculated from the current position of the vehicle and the sub-target. Sub-target angles contain more information than navigation commands represented by one-hot vectors. The sub-target angle is not only used to extract useful features, but also to assist the prediction layer to predict steering angle and throttle state (control acceleration). However, most existing solutions only consider visual camera frames as input, thus ignoring the temporal relationship between frames. In response to this situation, Eraqi et al. \cite{eraqi2017end} proposed a convolutional long short-term memory recurrent neural network (C-LSTM) that can be trained end-to-end to learn visual and dynamic driving temporal dependencies while introducing steering. The angular regression problem as classification strengthens the spatial relationship between neurons in the output layer. Although the prediction of steering angle has achieved good results in a single task, some researchers have found that it is effective to predict steering angle and speed simultaneously. Yang et al. \cite{yang2018end} proposed a multi-task learning framework to simultaneously predict cornering and speed control in an end-to-end manner. A network is proposed to predict discrete velocity commands and steering angles for image sequences. Taking the speed and visual information fed back by the underlying sensors as input, a multi-modal multitasking network is used to predict the speed value and steering angle. Huang et al. \cite{2021Multi} designed and compared various end-to-end multi-task deep learning networks using a combination of deep convolutional neural networks and long-short-term memory recurrent neural networks (CNN-LSTM), which can not only obtain the visual space in the driving scene Information, dynamic timing information can also be obtained, improving steering angle and speed prediction.

\subsubsection{Multiple Auxiliary Tasks}

When the vehicle is driving, visual input alone is not enough to make accurate steering judgments. By adding vehicle kinematics information, the behavior of the vehicle can be better estimated. The kinematic information ensures that the car does not perform driving behaviors that violate certain physical rules. Presumably, U-turns at 10 mph and 30 mph are different in terms of turning angle and control strategy. However, the visual observations given are almost the same.  While we can infer the speed of the vehicle from the speed of the scene change, it is still ambiguous and not easy to learn from images. The vehicle's own sensors can provide information such as current vehicle speed and steering. 

In addition, some auxiliary tasks applied in vehicle information fusion can also help us understand the surrounding environment of the vehicle. Under normal circumstances, drivers can quickly make driving decisions by paying attention to important information and understanding driving scenarios. Trained end-to-end autonomous driving with deep learning enables drivers to make corresponding judgments in driving scenarios by using supervised learning with driving behavior labels such as steering angle and speed. But in this method, it is difficult for the end-to-end model to learn and understand the mapping relationship between some important feature information and driving behavior. Areas that have a significant impact on driving decisions, such as vehicles, pedestrians, traffic lights, and drivable areas, are not given more attention. Therefore, multi-task-based learning is used in conjunction with multiple auxiliary tasks such as semantic segmentation and object detection, which helps to focus on salient regions and understand driving scenarios. 

Image segmentation \cite{badrinarayanan2017segnet} helps to better understand the environment. In autonomous driving, image segmentation is usually used to classify and understand The surrounding environment of a vehicle, which contains a lot of information, such as classifying surrounding vehicles,  pedestrians, road boundaries, buildings, etc. 

A segmented map can clearly identify the location of road boundaries and the location of surrounding vehicles on the road. This makes it easier for the vehicle to understand its stop/go or steering behavior. It significantly reduces the difficulty of learning everything implicitly from the original preprocessed images. Optical flow \cite{2015Optical} can identify the motion of objects. Transfer learning \cite{weiss2016survey} exploits the ability to share common features among tasks, and LSTM \cite{greff2016lstm} can be used to extract temporal information. Combining them into a single network and workflow can make up for the inadequacies in the single-camera perception process. Chen designed a new network structure, called Auxiliary Task Network (ATN), using auxiliary information other than the original image, which introduced human prior knowledge into the vehicle navigation. Image semantic segmentation, as an auxiliary task for navigation, introduces LSTM module and optical flow into the network to consider temporal information. To help improve driving performance while maintaining the benefits of minimal training data and end-to-end training methods.

\subsubsection{Lane Change prediction}

Another major scenario for using vehicle steering information is lane change. A German study reported that the probability of lane change on urban roads is 55\%, and the use rate of turn signals on highways is 75\%, considering other lane changes. Early predictors. Fusion of data from three sources usually yields the best prediction rates: 1) driver behavior observations (e.g. eye tracking) 2) sensor information about the environment (e.g. front/side radar, lane detection, GPS/map data) 3) vehicle parameters (e.g., turn signals, speed, acceleration, steering wheel angle). Steering wheel angle, as a directly measurable vehicle parameter, appears to be a promising early predictor of lane change. A mathematical model of steering wheel angle is proposed that helps predict lane changing maneuvers. The work \cite{zhu2020multimodal} in concentrated on the lane change intention prediction according to the sensory data, which contains the lane information given by a lane tracker, the vehicle velocity, lateral position and its derivation, and the steering wheel angle. Vehicle dynamic information is the direct response to the control actions from the driver. Schmidt et al. \cite{schmidt2014mathematical} proposed a lane change intention recognition method based on the construction of an explicit mathematical model of the steering wheel. In \cite{chen2017end}, driver lane change/keep intention inference systems were proposed on a driving simulator with the collection of vehicle dynamic information. In \cite{liu2019driving}, an intention recognition method with artificial neural networks (ANN) was proposed. CAN bus data and driver gaze information was collected and fed into the ANN. However, since vehicle dynamic information is the response to the driving actions, they give delayed information compared with driver behavior data and traffic context information in the intention inference tasks. In general, vehicle dynamic information cannot provide advanced information for intention prediction. However, they still useful for the intention identification and can help to recognize the intent at an early stage after the intended maneuver has been initiated.  

\subsubsection{Perception under noisy image data}

\begin{figure*}[ht]
\includegraphics[scale=1]{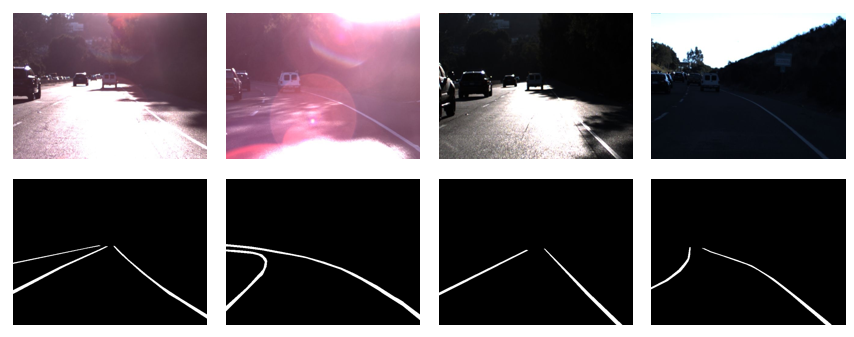}
\caption{Visual information is affected by noise. Fusion of the extracted lane line information and steering information to assist vehicle control or to perform lane line detection and segmentation} 
\label{fig2}
\end{figure*}

Many studies have shown that the perception of the vehicle will be greatly affected when the visual information is disturbed by noise. Tesla's driverless accident was caused by the failure of the perception module in a strong light environment \cite{eraqi2017end}. The low sun level during the tests caused significant problems in the operation of the models utilizing camera, as the wet road was easily overexposured in the camera image and there was even direct sunlight to the camera sensor, causing lens flares. These adverse lighting conditions have some variance during tests as the sun position changed during testing, which can affect the comparison of different
model results. As is shown in the Fig. \ref{fig2}

\subsection{Speed}
\subsubsection{Driving behavior classification}

At present, Drivers' abnormal driving behavior is the main cause of traffic accidents, such as drunk driving, fatigue driving, aggressive driving \cite{meiring2015review}. If there are some methods to predict the driver's driving behavior, it can remind the driver and reduce the occurrence of accidents. Some methods of monitoring driver behavior directly monitor the driver's face and body, but people can subjectively control their expressions, so this method may be deceptive and invade people's privacy. The driving behavior predicted by the underlying information of the vehicle cannot be subjectively changed by people, so the driving behavior predicted by the underlying information of the vehicle such as speed and acceleration is more accurate and convincing. Besides judging driving behavior can help prevent traffic accidents, it also provides some support for optimizing fuel consumption, because different driving behaviors have different fuel consumption. For example, aggressive driving often consumes more fuel than normal driving \cite{lee2011relationships}. At present, some insurance companies also have to judge the driver's usual driving behavior, so as to divide the insurance cost by different standards, if the driver is usually aggressive driving, then his cost will be higher than the average person \cite{weidner2017telematic}.

At the beginning of the development of using vehicle information to judge driving behavior, fewer signals were used. Sayed et al. \cite{sayed2001unobtrusive} proposed that only according to the vehicle's steering angle signal, based on the artificial neural network to determine whether the driver is sleepy, if the vehicle steering has not changed much, there is no some sharp change, then the driver is not awake. But in fact, there is no sufficient reason to think that the driver is sleepy. For example, the driving journey does not need a large steering, and the signal used in this method is single, and the accuracy is only 90\%, which needs to be further improved.

With the development of neural network and the research of judging driving behavior, there are more and more researches. Arefnezhad et al. \cite{arefnezhad2020applying} proposed method uses five signals, yaw rate, lateral acceleration, lateral deviation from the center of the road, steering wheel angle, and steering wheel angular velocity to determine whether the driver is awake, moderately sleepy, or extremely sleepy, and the drowsiness level is subdivided. Arefnezhad et al. \cite{arefnezhad2020applying} compared with Sayed et al. \cite{sayed2001unobtrusive}, they made full use of the vehicle information, judged from different angles, and considered it more comprehensively.

\begin{figure*}[ht]
\includegraphics[scale=0.52]{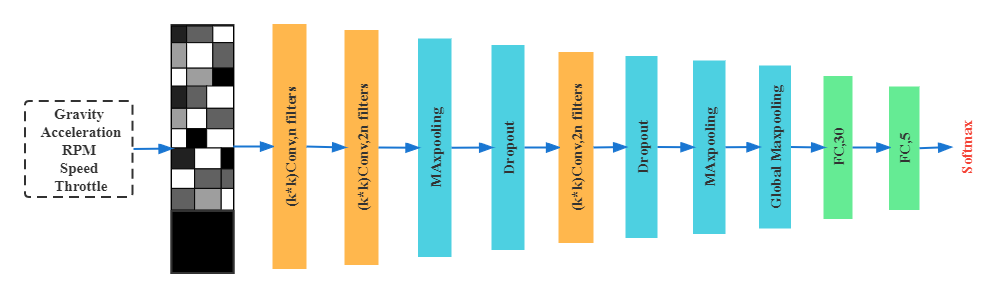}
\caption{Network structure for judging driving state based on recurrence plot technique.} 
\label{fig3}
\end{figure*}

Compared with the previous two methods, Shahverdy et al. \cite{shahverdy2020driver} were not limited to judging the driver's sleepy state, they further predicted whether the driver is normal, aggressive, distracted, drunk driving, and the behavior of judgment is more diverse. Most of the research is to input the vehicle information directly into the network, and is not related to the image, the novelty of this paper is that the acceleration, gravity, throttle, speed, and Revolutions Per Minute (RPM) information is converted into images through the recurrence plot technology, which can be applied to a variety of image-based neural networks. The network structure is shown in the Fig. \ref{fig3}.

When the driver performs some abnormal operations (such as sudden braking and acceleration), the acceleration and deceleration will change sharply, and the acceleration and deceleration change trends of different driving behaviors are different, so Jia et al. \cite{jia2020long} innovated from the extreme value of acceleration and deceleration of vehicle information, and judge different driving behaviors by detecting the extreme value of acceleration and deceleration.

In terms of the automatic transmission system of the car, if the gear is automatically changed according to the driver's driving intention, it can improve the efficiency of shifting, reduce fuel consumption, and save resources to protect the environment. Liu et al. \cite{liu2019driving} proposed the use of opening degree of the accelerator pedal, vehicle speed and brake pedal force to identify driver intent, and developed a shift strategy based on driver intention, which showed that this strategy can effectively reduce the number of shifts, thereby reducing fuel consumption.

Many researches were based on the hybrid network of 1D-CNN and LSTM or only on LSTM, while Cura et al. \cite{cura2020driver} found that the performance of only using 1D-CNN is equivalent to that of only using LSTM, and even the ability to distinguish aggressive driving is stronger.

\subsubsection{Vehicle speed prediction}
With the increase of population density, most of the air pollution in urban areas is caused by motor vehicle exhaust emissions. At present, many studies have optimized powertrain control by predicting vehicle status, so as to reduce fuel consumption, meet strict emission regulations, and reduce environmental pollution. In addition, in terms of autonomous driving, the biggest challenge facing cars is autonomous lateral and longitudinal control, including steering and speed control. Therefore, it is particularly important to accurately predict the speed.

In terms of driving habits, people usually slow down when turning corners and accelerate appropriately when going straight. Therefore, this behavior can continue to be imitated on autonomous vehicle, predicting whether the speed should increase or decrease according to the curvature of the road ahead \cite{sharma2019lateral}, so as to help the vehicle have better control.

Many methods of predicting vehicle speed use the information of the vehicle in a certain period of time to predict the speed in a certain period of time in the future. They only consider the information of the vehicle itself, and do not consider the factors that affect the speed control on the side such as the road traffic conditions and the distance from the surrounding vehicles. The influence of this external environment increases the pressure on the speed prediction. Jiang et al. \cite{jiang2016vehicle} first used neural network model to get the average traffic speed of the road section based on the previous traffic data, and then used Hidden Markov Model (HMM) to establish the statistical relationship between individual vehicle speed and average traffic speed, so as to predict the speed. Yeon et al. \cite{yeon2019ego} broke the limitation of predicting vehicle speed under specific conditions. They not only used the internal information of the vehicle such as speed, acceleration and engine speed to predict the vehicle speed, but also used the information such as the relative speed and distance with the vehicle in front, as well as the position of the vehicle. Through the internal and external information in 30 seconds, they jointly predicted the vehicle speed in the next 15 seconds. These methods fully consider the external environment information, and consider the factors that affect the speed prediction more comprehensively.

What data to use to predict the speed needs sufficient reasons. When too little data is used, it is not enough to reflect the vehicle state. When too much data is used, irrelevant data will interfere with the model. Xing et al. \cite{xing2021dual} collected a variety of experimental data, mining the deep value of the data, such as the VCU speed and VBox speed, so that the measured speed is more accurate, and Xing et al. \cite{xing2021dual} added the information of the opening of the driving pedal and the opening of the brake pedal to the model, fully considering the intention of the driver to accelerate and decelerate. Xing et al. \cite{xing2021energy} also made a full selection on the input signal. In most studies, people choose the input signal to predict the vehicle speed according to their own feelings, without too much consideration on the quality of the signal, which is subjective. Xing et al. \cite{xing2021energy} used Pearson correlation to judge the relationship between vehicle signals and vehicle speed, and selected the 15 signals with the highest scores as the input of the model, ensuring the data quality and making the predicted vehicle speed more accurate to a certain extent.

Most of the above methods are based on fuel vehicles. In terms of electric vehicles, electric vehicles have gradually entered people's lives because of their advantages such as no pollution, low noise and no need for frequent maintenance. Zhang et al. \cite{zhang2020improved} 
\cite{zhang2021real} broke the limitation of using only vehicle information to predict vehicle speed, used image and vehicle speed fusion to jointly predict vehicle speed, and further optimized the energy management strategy of electric vehicles on this basis.

Most studies on speed prediction are carried out under normal geographical conditions. Affected by air pressure, temperature and humidity, oxygen content, etc., different altitudes have different effects on vehicle dynamic performance, and people's driving habits are also different. Therefore, specific models are needed to predict vehicle speed in high altitude areas. Shi et al. \cite{shi2019short} considered the influence of altitude on the effect of speed prediction model, and established a speed prediction neural network model in high altitude areas.

\begin{figure*}[ht]
\includegraphics[scale=0.58]{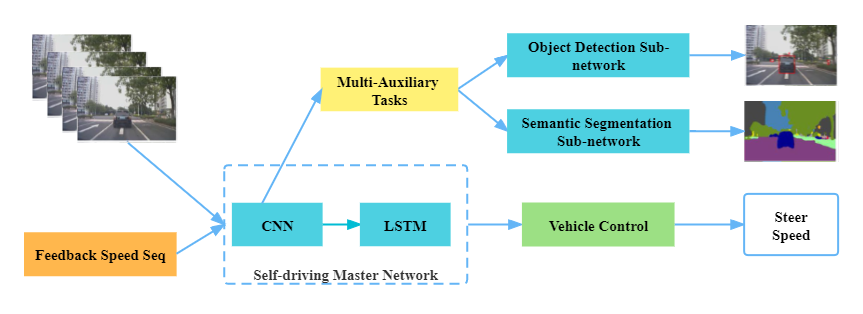}
\captionsetup{singlelinecheck=off}
\caption{End-to-end network 
architecture with auxiliary task.} 
\label{fig4}
\end{figure*}

Many researches based on vehicle bottom information are not connected with perception tasks. Wang et al. \cite{wang2019end} not only predicted steering angle and speed at the same time, but also shared the predicted network on semantic segmentation and target detection tasks. It can efficiently carry out three tasks at the same time, making a breakthrough in bottom information and perception tasks. The network structure is shown in the Fig. \ref{fig4}.

\subsubsection{Trajectory prediction}
If the trajectory prediction of vehicles on the road is carried out in real time on the traffic monitoring system, or a set of trajectory prediction system is applied to vehicles, then when the trajectory between vehicles will conflict, the trajectory prediction system can reduce a large number of traffic accidents by giving owners a certain degree of reminder.

On the highway, the speed of vehicle driving is faster, the lane change process is more dangerous, if you can predict in time whether the vehicle is in a safe state during the lane change, you can affect the driver's behavior and avoid accidents. Tomar et al. \cite{tomar2010prediction} proposed a lane change model, which predicts the next state during lane changing through the relative speed between vehicles, safe distance, whether the vehicle is accelerating or decelerating, and the current vehicle state (collison, near collison, absolute safe and safe), so as to ensure the safety during lane changing.

Traditional prediction trajectory methods use physics-based motion models, and while their results are accurate in the short term, if they exceed one second, the results are less reliable and cannot be further referenced. Therefore, Jeong et al. \cite{jeong2017long} used the advantages of neural network to fuse the underlying information such as vehicle speed, acceleration, steering and road conditions, and predicted the position of the vehicle in 0-4 seconds, which provided a reference for auto drive system and road safety.

\subsubsection{Mileage, time, energy consumption forecasts}
Although the fuel consumption meter of the car can now display the instantaneous fuel consumption, average fuel consumption, and mileage of the car, the mileage of the car is calculated based on the remaining fuel in the car fuel tank divided by the average fuel consumption of the car. The average fuel consumption is calculated according to the mileage and fuel consumed over a period of time, the real-time is not high. Moreover, if both the remaining time and the remaining mileage can be displayed on the vehicle display screen according to the real-time information, it is more informative, so that the driver can arrange the itinerary according to a variety of information. Therefore, using parameters that affect fuel usage such as vehicle speed and road slope to predict the remaining mileage and time can provide greater help to the driver. Estimating the real-time energy consumption of vehicles allows for monitoring road pollution and 
providing data support for pollution reduction technology.

In terms of fuel vehicles,  Chen et al. \cite{chen2012artificial} proposed a method of predicting the remaining mileage and time of the vehicle based on the comprehensive vehicle and road information, not only based on the fuel capacity of the vehicle itself, the current speed, the engine speed, the weight of the vehicle and other information, but also based on the road slope and other road information, through multi-faceted information prediction is more real-time and accurate than only the reference to the average fuel consumption. In terms of electric vehicles, the real-time energy consumption of electric vehicles is estimated by vehicle speed, road altitude, etc., and the proposed model has the lowest error compared with existing technologies \cite{modi2020estimation}.

If you want to reduce the pollution of road traffic, you must not only have effective environmental optimization technology, but also have real-time and convenient monitoring methods, and accurate monitoring can be better optimized. Kanarachos et al. \cite{kanarachos2019instantaneous} proposed a method that can estimate vehicle energy consumption through neural network only by using the GPS location, speed, altitude and other information of the smart phone, without the need to use complex and cumbersome monitoring methods, which is accurate and simple, and brings new methods to monitor fuel consumption.


\subsubsection{Sideslip angle, Vehicle deceleration prediction}

In the case of understeering or oversteering (such as drift), the vehicle's stability control system is particularly important, and the stability control system restores the vehicle to a stable state by interfering with the engine or wheels. In terms of lateral stability control, the slip angle and roll angle are the key parameters of the lateral stability of the vehicle, but these parameters require very expensive equipment measurement, Melzi et al. \cite{melzi2011vehicle} based on the vehicle's lateral acceleration, yaw rate, speed, and steering angle, the lateral slip angle is estimated by the neural network, which greatly reduces the cost and the accuracy of the method is proved to be high \cite{gonzalez2020simultaneous}. The roll angle and the slip angle are estimated at the same time, and the cost is greatly reduced.

If the driver is given a certain amount of warning before a collision occurs, a large number of traffic accidents will be reduced, and even in the case of automatic driving, it also takes a certain period of time to respond. Brake deceleration refers to the ability of a vehicle to rapidly reduce its travel speed until it stops while on the move. Zhang et al. \cite{zhang2022prediction} proposed that according to the relative distance between vehicles, the relative speed of vehicles, the acceleration of the vehicle in front of the vehicle and the lateral distance, the braking deceleration of the vehicle should be effectively predicted, and the ability of the vehicle to immediately slow down and stop should be judged.

\section{Fusion Methodology}
\label{3}
For multi-modal fusion, it is extremely important to decide what information fusion and fusion method to use. The recurrence plot and spectrogram, which can transform one-dimensional information into two-dimensional images and are convenient for two-dimensional convolutional neural networks, are first introduced in detail. Then we present efficient tensor fusion and adaptive multi-modal fusion technology.
\subsection{Recurrence Plot}
Recurrence Plot(RP) was first proposed in 1987 for qualitative analysis of nonlinear dynamical systems. A recurrence plot is an image obtained from time series data representing the distance between each point in time, and the image can be binarized using thresholds. General recursive graphs are an important method for analyzing the periodicity, chaos, and non-stationary nature of time series, revealing the internal structure of time series, giving a priori knowledge about similarity, amount of information, and predictability, and recurrence plot is particularly suitable for short time series data. The RP has been used recently to identify changes of dynamic patterns in time series in many other fields, for instance in financial data time series \cite{belaire2002assessing,strozzi2007recurrence} and ecosystem time series \cite{facchini2007nonlinear}
. Because RP can convert time series into two-dimensional information, it is especially suitable for two-dimensional convolutional neural networks, which provides important support for the transformation of one-dimensional information and two-dimensional information. Shahverdy et al. \cite{shahverdy2020driver} used this technology to convert the vehicle bottom information into recurrence plot, and then input it into convolutional neural network. 

\subsection{Spectrogram}
In the field of signal processing, there are three main domain angles to analyze the signal, namely the time domain, the frequency domain, and the time frequency domain, corresponding to the time domain map, the frequency domain map, and the time-frequency map, that is, the language spectrum. Both the time and frequency domains can represent information in only two dimensions of a signal, while a spectrogram uses two-dimensional images to represent information in three dimensions. The abscissa of the spectrogram is time, the ordinate is the frequency, and the coordinate point value is the speech data energy. Since the two-dimensional plane is used to express three-dimensional information, the size of the energy value is expressed by color, and the color depth indicates that the stronger the speech energy of the point \cite{wyse2017audio}, which is a comprehensive description of the audio in the time domain and frequency domain characteristics. Spectrogram is based on the short-term Fourier transform(STFT) and is very helpful in analyzing the time-frequency characteristics of a signal. STFT is the most classic time-frequency domain analysis method, STFT by a long period of signal framing, windowing, and then do Fourier transform (FFT) for each frame, and finally stack the results of each frame along another dimension, forming a Spectrum. Arandjelovic et al. \cite{arandjelovic2017look} \cite{arandjelovic2018objects} converted  the sound into a log-spectrum, and then through a series of convolutional neural network processing, to determine whether the video matches the sound and locate the
sounding object.

The steering angle, vehicle speed, acceleration and other information of the vehicle change with time. The spectrum diagram is an image that can reflect the change of frequency with time. It can convert one-dimensional information into two-dimensional information, so it can be more conveniently applied to convolutional neural network.

\begin{figure}[ht]
\centering 
\includegraphics[scale=0.42]{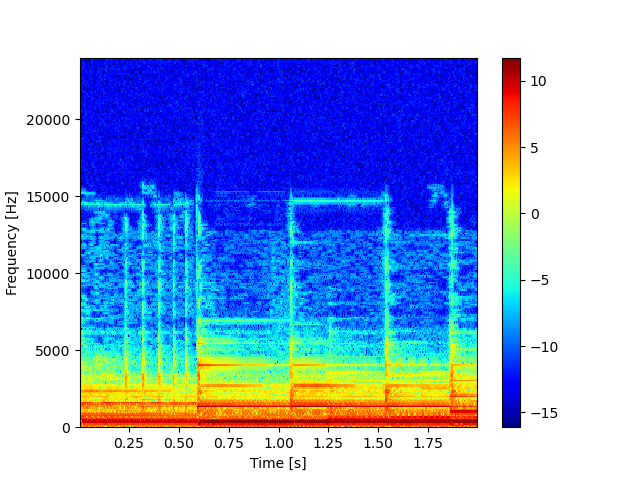}

\captionsetup[figure]{name={Fig.},labelsep=period,singlelinecheck=off} 

\caption{Abscissa is time, ordinate is frequency, and coordinate point value is voice data energy. The size of energy value is indicated by color. The darker the color, the stronger the speech energy at this point.}

\label{spectrogram}
\end{figure}
\subsection{Tensor fusion}
As a mainstream fusion method of multi-modal information fusion, tensor fusion is widely used in various fields, and the field of automatic driving is no exception. There are many ways of tensor fusion, such as early simple feature stitching, late decision fusion, tensor outer product fusion and so on. In this paper, we want to introduce a widely used and novel fusion method - tensor outer product.

As can be seen from the above, in the field of automatic driving, in addition to image and other data information, the data information at the bottom of the vehicle, such as steering angle and vehicle speed, is also very important. Through the tensor outer product method, various such vehicle bottom information and image data information can be fused,  to better solve practical problems. This fusion method can fuse the features between various data information more fully and flexibly, and the effect is better than simple feature stitching and late decision-making.

Zadeh et al. \cite{zadeh2017tensor} proposed Tensor Fusion Network based on tensor outer product fusion, and many subsequent optimizations and variant models are also based on this network model. The tensor fusion method used in the network is the tensor outer product, which is explained in detail. Tensor outer product is the outer product operation of the feature vector extracted from each mode to obtain a high-dimensional fusion tensor Z, and then project the high-dimensional fusion tensor Z into the low-dimensional space through a linear layer. Each element of each eigenvector is fully fused. After the fusion of two modes, a second-order tensor is formed, and after the fusion of three modes, a third-order tensor is formed. Several figures show the process of a tensor outer product.

The Tensor Fusion Network  needs external product operation for the feature vectors of each mode, when there are many feature vectors to be fused, the network will carry out the high-dimensional tensor calculation, and the calculation cost will be very high. For example, when we need to fuse the three information of steering angle, vehicle speed, and image at the same time, after fusion, we will get a third-order tensor Z, If we want to use the linear layer to project it into the low dimensional space, we need a fourth-order weight matrix W and Z to complete the calculation.

In view of the high computational cost of Tensor Fusion Network, many network models have proposed different solutions: Liu et al. \cite{liu2018efficient} were also based on the operation of the tensor outer product. Through rigorous mathematical derivation, this paper decomposes the parameter W and fusion tensor Z the of Tensor Fusion Network. Finally, the high-order tensor operation is decomposed into linear operation, so that the calculation cost will not increase exponentially with the increase of mode. Based on this, a Low-rank Multi-modal Fusion is proposed. The article uses Fourier convolution to replace the previous high-order tensor calculation, so as to solve the problem of high computational cost, and proposes a Multi-modal Compact Bilinear pooling (MCB) fusion network. The hierarchical plural fusion network (HPFN) proposed in the article \cite{hou2019deep} not only adopts Low-rank Multi-modal Fusion strategy similar to the article \cite{liu2018efficient} to solve the problem of high computational cost, but also realizes local feature fusion through the added sliding window mechanism The feature fusion of multiple time periods and the fusion sequence can be controlled to make the feature fusion more sufficient; Zhu et al. \cite{zhu2020multimodal} improved the Low-rank Multi-modal Fusion network  and solved the problem that the Tensor Fusion Network ignores the correlation between various modes by adding self attention mechanism. The specific operation is to change each mode into a new single-mode feature vector through the self attention mechanism module, and then fuse through Low-rank Multi-modal Fusion.

\subsection{Adaptive fusion}
Compared with the above fusion methods, adaptive fusion is more flexible and natural, because the network using this fusion method will not determine a specific fusion operation, such as feature stitching, tensor outer product, etc., but let the network decide "how" to more effectively integrate a given set of multi-modal features. Sahu et al. \cite{sahu2019adaptive} proposed two adaptive fusion network structures: 1) Auto-Fusion, which encodes the information of all modes and splices them into a tensor, then restores the features with the decoder, and finally calculates the loss between the features. This method not only integrates the feature vectors, but also learns the useful features. It solves the problem that the final predictor bears the additional responsibility of identifying useful signals. 2) GAN-Fusion, the network first finds a trunk mode, then fuses the other modal information except for the trunk mode, and fuses the fused information against the information of the trunk mode, so as to obtain the new feature vector of the trunk mode. In the same operation, all modes are used as the trunk mode at the same time, so we can obtain the new feature vector of each mode, and then splice these feature vectors. That is to complete the final integration.

\section{Datasets}
\label{4}
Many datasets only contain image and lidar information for image segmentation, target detection and other tasks, but few datasets contain a large amount of vehicle information at the same time. This chapter shows some datasets covering images, lidar, vehicle information and so on, which makes the research of multi-modal fusion based on vehicle information more convenient.

\begin{figure*}[!htb]
\includegraphics[width=\textwidth]{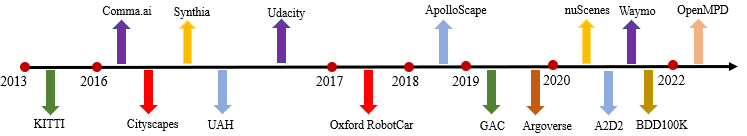}
\caption{Dataset timeline with vehicle underlying information} \label{fig5}
\end{figure*}

\begin{table*}
\centering
\def\arraystretch{1.5}
\addtolength{\leftskip}{-3cm}
\addtolength{\rightskip}{-3cm}
\captionsetup[table]{
  labelsep=newline,
  singlelinecheck=false,
}
\caption{\label{tab:test}Public datasets for autonomous driving. We describe in detail the number of images contained in the data set and whether it contains lidar data, radar data, GPS data, vehicle bus data and control related data.}
\begin{tabular}{lccccccl}
\hline
Dataset & Samples  & Image Type & LIDAR & RADAR  & GPS & Bus Data & Control Actions \\
\hline
Udacity& $34,000$   & $RGB$  & $YES$   & $NO$ & $YES$  &$YES$& $Steering wheel,Speed$\\
Waymo \cite{sun2020scalability}& $1,000,000$   & $RGB$  & $YES$   & $NO$ & $YES$  &$NO$& $Steering wheel$\\
Comma.ai \cite{santana2016learning} & $video$ & $RGB$ & $NO$  & $NO$ & $YES$ & $YES$& $Steering wheel,Speed$\\
GAC \cite{wang2019end}& $3,240,000$   & $RGB$  & $NO$   & $NO$ & $-$  &$-$& $Steering wheel,Speed$\\
A2D2 \cite{geyer2020a2d2}& $433,833$   & $RGB$  & $YES$   & $NO$ & $YES$  &$YES$ & $Steering wheel,Speed,Acceleration$\\
Cityscapes \cite{cordts2016cityscapes}& $25,000$ &$RGB$  & $NO$  & $NO$ & $YES$ & $YES$ &$-$\\
KITTI \cite{geiger2013vision}& $14,999$ &$RGB$  & $YES$  & $NO$ & $YES$ & $NO$ &$-$\\
Synthia \cite{ros2016synthia} & $213,400$ & $RGB$ & $YES$  & $NO$ & $NO$ & $-$&$-$\\
Oxford RobotCar \cite{maddern20171} & $20,000,000$ & $RGB$ & $YES$  & $YES$ & $YES$ & $YES$&$Speed$\\
nuScenes \cite{caesar2020nuscenes} & $1,400,000$ & $RGB$ & $YES$  & $YES$ & $YES$ & $YES$&$Steering wheel,Speed,Acceleration$\\
BDD100K \cite{yu2020bdd100k} & $120,000,000$ & $RGB$ & $NO$  & $NO$ & $YES$ & $NO$&$-$\\
ApolloScape \cite{huang2018apolloscape} & $143,906$ & $RGB$ & $YES$  & $NO$ & $YES$ & $YES$&$-$\\
Argoverse \cite{chang2019argoverse} & $video$ & $RGB$ & $YES$  & $NO$ & $YES$ & $YES$&$-$\\
UAH \cite{romera2016need} & $video$ & $RGB$ & $NO$  & $NO$ & $YES$ & $YES$&$Steering wheel,Speed,Acceleration$\\
OpenMPD \cite{zhang2022openmpd} & $15,000$ & $RGB$ & $YES$  & $NO$ & $NO$ & $YES$&$Steering wheel,Speed,Acceleration$\\
\hline
\end{tabular}
\end{table*}

\subsection{UAH-DriveSet}  
UAH \cite{romera2016need} is a public driving analysis data set released in 2016. Six drivers of different ages and genders drive different vehicles on different roads (motorway and secondary road) to simulate three different driving behaviors (normal, drowsy and aggressive), so as to collect data, including video, original vehicle data and processed data. The UAH-DriveSet is available at: http:// www.robesafe.com/personal/eduardo.romera/uah-driveset.

Each folder of the dataset includes video and data text files. Data files include: measurements obtained by the mobile phone sensor; variables that are processed in real-time by DriveSafe App in a  continuous way((car position relative to lane center (meters), car angle relative to lane curvature (degrees), Distance to ahead vehicle in current lane (meters), etc)); a single event generated during driving detected by the DriveSafe(brake, accelerate, steer, etc); Drivesafe's rating of driver behavior. Vehicle bus data mainly includes vehicle speed, acceleration, longitude and latitude, etc.

The dataset also provides a reader tool, by selecting a route that can be synchronized in the interface to display video and recorded data changes in real-time, which can facilitate researchers to observe the performance of vehicles in a certain driving style, providing researchers with good analysis tools.

\subsection{BDD100K}
The BDD100K dataset \cite{yu2020bdd100k}, released in 2018 by the University of Berkeley AI Lab (BAIR), is a classic large-scale, diverse dataset of driving videos.

The dataset includes image information, as well as GPS/IMU information to save driving trajectories, with images characterized by high resolution (720p) and high frame rate (30fps). The dataset contains a total of 100,000 driving videos (40 seconds each) in different areas (New York, San Francisco Bay Area, and other areas), different scenarios (city streets, residential areas, and highways), different weather, and different times. Due to the diversity of its external environment, the robustness of the applied model is improved. Dataset videos are divided into training (70K), validation (10K), and testing (20K) sets. The 10th second frame in each video is annotated for the image task, and the entire sequence is used for the tracking task. The dataset can be used for tasks such as object detection, lane line detection, semantic instance segmentation, multiple object tracking, and segmentation.

\subsection{A2D2}
Audi Autonomous Driving Dataset (A2D2) \cite{geyer2020a2d2} is a public driving dataset released by Audi in 2020. Data and other details are available from the http://www.a2d2.audi. The A2D2 dataset also has a certain diversity, taking into account not only the diversity of road environments (highways, country roads and urban roads in southern Germany), but also the diversity of weather (cloudy, rainy and sunny days). Datasets include image data and 3D point clouds, as well as 3D bounding boxes, semantic segmentation, instance segmentation, and vehicle underlying data extracted from the car bus, which can be used for a wide variety of tasks due to the diversity of their annotations. Vehicle bus data mainly includes steering angle, brake, throttle, odometry, etc. The dataset included 41227 frames with semantic segmentation labels and point cloud labels, of which 12497 frames contained 3D callout boxes for targets within the field of view of the front camera, and also included 392,556 consecutive frames of unlabeled sensor data.

\subsection{Udacity}

In collecting vehicle dynamics information, ROS provides a flexible ecosystem and includes powerful tools and libraries. Many researchers use the ROS framework for dataset creation. ROS nodes can perform synchronous data collection from the CAN bus and RGB cameras. Through the ROS node information, the dynamic information of the vehicle during the driving process, such as speed, steering angle, braking, torque, and accelerator value, can be obtained.

Udacity's autonomous driving dataset, 1920x1200 resolution images captured using a Point Grey research camera, the data collected is divided into two datasets: the first consists of daytime conditions in Mountain View, California and adjacent cities The dataset contains more than 65,000 annotated objects in 9,423 frames, and the annotation method combines machine and human. The labels are cars, trucks, and pedestrians; the second dataset is generally similar to the former, except that the annotation content of traffic lights is increased, the number of datasets has also increased to 15,000 frames, and the annotation method is completely manual. In addition to the images taken by the vehicle, the content of the dataset also includes the attributes and parameters of the vehicle itself, such as latitude and longitude, brakes, accelerator, steering, and rotational speed. Many researchers have performed the above fusion research on the vehicle's underlying information and visual data on the Udacity dataset. 

\subsection{Comma.ai}
The Comma.ai Dataset is a video dataset for autonomous driving containing a total of 7.25 hours of video, which contains 10 videos recorded at 20Hz by a video camera mounted on an Acura ILX 2016 windshield Camera recording. In addition to this, the dataset also contains some measurements of the vehicle itself, such as car speed, acceleration, steering angle, GPS coordinates, gyroscope angle, etc. The measurements are all converted to a uniform 100Hz time base, the Comma.ai dataset is provided by Comma.ai. ai company was released in 2016.

\subsection{OpenMPD}
OpenMPD is a multi-modal autonomous driving dataset we released in 2022. Compared with the existing dataset, OpenMPD includes more complex driving scenes (such as night, road construction, U-turn, etc.) and complex roads (such as intersections, tunnels/culvert, viaducts, etc.) to provide diverse data for autonomous driving tasks. Our vehicle is equipped with six cameras and four LiDARs to achieve 360-degree full coverage while acquiring multi-modal data. In particular, we use 128-beam LiDAR to collect high-resolution dense point cloud data. We have extracted 15K keyframes for annotation, including 2D/3D semantic segmentation, 2D/3D bounding box, which can be used for semantic segmentation, object detection, object tracking and other multi-task, multi-view research. We also collect vehicle underlying information such as vehicle speed, acceleration, steering angle and engine speed from CAN bus. For data and more information, please visit http://www.openmpd.com/.

\section{Challenges} 
\label{5}

Most current autonomous driving methods focus on how to directly learn end-to-end autonomous driving control policies from raw sensory data. Essentially, this control strategy can be viewed as a mapping between images and driving behavior, which usually suffers from low generalization ability. Although much research has been done on the joint learning of vehicle information and images in this paper, there are still many challenges to be further investigated. In this section, we will discuss some open problems from vehicle control, scene generalization, and datasets, as follows.

The automatic driving function of the vehicle is mainly realized by the longitudinal motion control and the lateral motion control. In automatic vehicle control, longitudinal control is still a challenging problem \cite{sharma2018behavioral}. In addition to the visual information collected by camera radar, with the increase of the number of on-board sensors, more and more vehicle motion state parameters can be collected. For example, longitudinal acceleration, air resistance, tire load, ground friction, ground inclination, etc. By combining image perception information, vehicles can achieve better longitudinal control in constant speed cruise, adaptive cruise and anti-collision systems. Therefore, researchers can try to combine rich self sensor information from different driving scenes to achieve vehicle control.

In terms of research scenarios, most studies currently focus on daytime driving scenarios. In works such as \cite{bojarski2016end} and \cite{wang2019end}, only a few of their studies involve nighttime driving, while as stated in \cite{chen2017end}, most The researchers ignored nighttime driving in their scenarios. If automatic driving only relies on daytime driving, its range of motion is limited. However, the information of self owned sensors is not affected by bad weather and light conditions. Therefore, we suggest that researchers can widely use vehicle information in future research to improve scene generalization ability.

In terms of datasets, there are relatively few publicly available datasets with vehicle underlying sensor information. Most researchers collect fixed data for specific scenarios or conditions, and large-scale autonomous driving datasets with vehicle kinematic parameters are lacking. A common solution is to perform data augmentation on the limited data to obtain additional training data. Such as common flip, crop, increase shadow operations. Or convert the color, brightness, intensity and space of the image. However, it still has certain limitations compared to real datasets covering various lighting conditions and complex road conditions. In addition, many researchers have built autonomous driving simulation scenarios on simulators and obtained simulated data for model training, but the error has almost doubled from the simulation to the real world. The complexity and diversity of real-world environments pose even greater challenges. This suggests the need for training and testing on larger real-world datasets and more realistic simulated environments in the future.

\begin{table*}[ht]
\centering
\caption{On the UAH dataset, the performance of our method of fusing vehicle speed and image is compared with other driving behavior classification methods. The Acc, Pre and Rec represent the accuracy, precision and recall. The "-" means that it is not indicated in the method.}
\def\arraystretch{1.3}
\setlength{\tabcolsep}{5mm}{
\begin{tabular}{c|c|cccc}
\hline
Type                               & Model                                           & F1       & Acc             & Pre      & Rec         \\ \hline
\multirow{13}{*}{Machine Learning} 
                                   & MLP \cite{saleh2017driving}                            & 48.0\%           & -                  & -           & -           \\
                                   & Decision Tree \cite{saleh2017driving}                            & 80.0\%           & -                  & -           & - 
                \\
                                 & Generic Model \cite{yi2019machine}                            & -          & 81.3\%                 & 82.2\%          & 80.7\%          \\
                                   & Personalized Model \cite{yi2019machine}                        & -          & 89.5\%                 & 89.9\%          & 89.2\%          \\
                                   & Road-Aware Model \cite{yi2019machine}                          & -          & 83.6\%                 & 84.4\%          & 83.2\%          \\
                                    & Personalized Model With Road Information \cite{yi2019machine} & - & 91.6\% & 92.0\% & 91.5\%          \\
                                   & Discriminant Analy \cite{yi2019machine}                        & -              & 51.1\%                 & -              & -              \\
                                   & Decision ree \cite{yi2019machine}                              & -              & 63.1\%                 & -              & -              \\
                                   & KNN \cite{yi2019machine}                                      & -              & 66.8\%                 & -              & -              \\
                                   & SVM \cite{yi2019machine}                                       & -              & 67.4\%                 & -              & -              \\
                                   & Ensemble Learning \cite{yi2019machine}                         & -              & 81.3\%                 & -              & -              \\
                                   & DriveBFR \cite{vyas2021drivebfr}                                 & 95.0\%           & -                     & -              & -              \\ \hline
\multirow{4}{*}{Neural Netwowk}    & Stacked-LSTM \cite{saleh2017driving}                             & 91.0\%           & -                     & -              & -              \\
                                   & HDC-ANN \cite{schlegel2021multivariate}                                  & 94.0\%           & -                     & -              & -              \\
                                   & HDC-SNN \cite{schlegel2021multivariate}                                  & 92.0\%           & -                     & -              & -              \\
                                 
                                  & Ours                                             & \textbf{97.1\%} & \textbf{97.3\%}         & \textbf{97.0\%} & \textbf{97.3\%} \\ \hline
\end{tabular}}
\label{tab:all}
\end{table*}

\begin{table}[ht]
\caption{Performance comparison with other driving behavior classification methods on motorway road.}
\def\arraystretch{1.3}
\setlength{\tabcolsep}{0.65mm}{
\begin{tabular}{c|c|cccc}
\hline
Type                              & Model                      & F1       & Acc      & Pre      & Rec         \\ \hline
\multirow{5}{*}{Machine Learning} & Logistic Regression \cite{ghandour2021driver} & 53.0\%           & 54.0\%           & 53.0\%           & 54.0\%           \\
                                  & Gradient Boosting \cite{ghandour2021driver} & 68.0\%           & 67.0\%           & 70.0\%           & 67.0\%           \\
                                  & Random Forest \cite{ghandour2021driver}       & 63.0\%           & 63.0\%           & 63.0\%           & 63.0\%           \\
                                  & MLP \cite{saleh2017driving}                 & 51.0\%           & -              & -              & -              \\
                                  & Decision-Tree \cite{saleh2017driving}       & 75.0\%           & -              & -              & -              \\ \hline
\multirow{3}{*}{Neural Network}   & Neural Network \cite{ghandour2021driver}      & 25.0\%           & 29.0\%           & 29.0\%           & 27.0\%           \\
                                  & Stacked-LSTM \cite{saleh2017driving}        & 86.0\%           & -              & -              & -              \\
                                  
                                  & Our                        & \textbf{93.4\%} & \textbf{93.5\%} & \textbf{93.5\%} & \textbf{93.5\%} \\ \hline
\end{tabular}}
\label{tab:motor}
\end{table}

\begin{table}[ht]
\caption{Performance comparison with other driving behavior classification methods on secondary road.}
\def\arraystretch{1.3}
\setlength{\tabcolsep}{0.7mm}{
\begin{tabular}{c|c|cccc}
\hline
Type                              & Model                      & F1       & Acc       & Pre      & Rec        \\ \hline
\multirow{5}{*}{Machine Learning} & Logistic Regression\cite{ghandour2021machine} & 51.0\%           & 55.0\%           & 54.0\%           & 52.0\%           \\
                                  & Random Forest \cite{ghandour2021machine}       & 61.0\%           & 63.0\%           & 63.0\%           & 62.0\%           \\
                                  & Gradient Boosting \cite{ghandour2021machine}   & 64.0\%           & 65.0\%           & 65.0\%           & 64.0\%           \\
                                  & MLP \cite{saleh2017driving}                 & 64.0\%           & -              & -              & -              \\
                                  & Decision-Tree \cite{saleh2017driving}       & 92.0\%           & -              & -              & -              \\ \hline
\multirow{2}{*}{Neural Network}   & Stacked-LSTM \cite{saleh2017driving}        & 95.0\%           & -              & -              & -              \\
                                  
                                  & Our                        & \textbf{98.3\%} & \textbf{98.4\%} & \textbf{98.3\%} & \textbf{98.3\%} \\ \hline
\end{tabular}}
\label{tab:sec}
\end{table}

\section{Future Research Prospects} 
\label{6}
Compared with additional sensors such as laser radar, vehicle information is not limited by the external environment and sensor hardware, and it is low-cost and can provide accurate and real-time data. However, many studies have not made full use of these advantages, especially in robot autonomous control, automatic driving and so on. The following describes our future research directions in terms of vehicle speed and steering angle.

\subsection{vehicle speed}

Accurately distinguishing the driving behavior of drivers plays an important role in driving assistance system, road safety, energy optimization and so on. The behavior classification method based on driver dynamics directly uses the camera to aim at the driver's face and body, which violates the driver's privacy. The behavior classification method based on vehicle dynamics only uses vehicle information for analysis, which lacks images that can provide rich information around vehicles and roads. The change degree of vehicle information directly reflects the driver's driving behavior. For example, the sharp change of speed in a short time reflects that the driver is in an aggressive state. The road image can reflect the degree of emptiness, congestion and the distance from obstacles. If the vehicle information is combined with the image, it can make the information complementary and improve the classification performance.

We have explored this aspect. 1D information can be converted into 2D information through Gramian Angular Field, Markov Transition Field, Recurrence Plot and Short-time Fourier Transform. We first convert the temporal speed data within five seconds into 2D, and then get their own features with the road end image through two feature extraction convolution networks, then fuse them through MSELoss. The MSELoss between the vehicle speed and the image features are calculated to force them to align. Finally, the MSELoss value is mapped to three driving behaviors through the full connection layer. We judge the driver's behavior(normal, drowsy, aggressive) according to the road(all road, motorway road, secondary road) in the public data set UAH, as shown in Table \ref{tab:all}, Table \ref{tab:motor} and Table \ref{tab:sec}. Table \ref{tab:all} shows our comparison with other methods on all roads, which is 2.1\% higher than the current best DriveBFR. Table \ref{tab:motor} shows our comparison with other methods on motorway road, which is 7.4\% higher than the current best Stacked-LSTM. Table \ref{tab:sec} shows our comparison with other methods on secondary road, which is 3.3\% higher than the best Stacked-LSTM. Experiments show that speed and image can be effectively fused to improve classification performance.

For autonomous vehicles, due to the fixed frequency of camera shooting, the vehicle can only track the target blindly without knowing the degree of object position change. The vehicle information such as vehicle speed intuitively reflects the change degree of the object position. The faster the speed between adjacent frames, the greater the change of the target position, and the lower the object similarity between images. Therefore, the vehicle speed can be used as auxiliary information of target tracking, which can be combined with the image to improve the target detection ability.

\subsection{steering angle}
Steering angle prediction can be incorporated into a wider range of computer vision-based techniques such as lane line detection, obstacle detection, etc. In the field of autonomous driving, one of the necessary capabilities of vehicles is the need to accurately identify lane lines. According to different road conditions, reasonable lane keeping to prevent deviation from the safe driving area is inseparable from accurate steering angle prediction. During the turning process, since the lane line is in a curved state and the lane line spans different areas of the image, the convolution kernel in the current convolutional neural network has limitations in extracting feature regions, and cannot efficiently extract curved and cross-regional features. The turning angle describes the size of the turning span, that is, the bending degree of the lane line, which intuitively reflects the state of the lane line. Therefore, the turning angle can be used as auxiliary data and images to improve the recognition ability of lane lines.

In order to prove the correctness of our theory, we have carried out a lot of work and experiments in this field. First of all, we extracted more than 6000 road end images and steering angle data corresponding to each image from Udacity, a public data set in the field of automatic driving, and annotated the data. Secondly, we used the classic lane line detection algorithm Lanenet \cite{neven2018towards} as our baseline.Based on this baseline, we integrate the steering angle information into the feature map obtained after the shared encoder. First, we perform maxpool operation on the feature map to obtain one-dimensional feature information about the feature map. At the same time, for the steering angle data, we extract the features through MLP, and then combine the one-dimensional feature information of the feature map with the steering angle feature vector after feature extraction to obtain the channel attention vector of the feature map, the vector is multiplied with the original feature map to obtain the new feature map that we fuse the steering angle. And classified it into two groups by light intensity and whether the lane line is curved, to observe the index changes after adding steering angle in different scenes, as shown in Table \ref{angle}. The experimental results show that the steering angle and image fusion can improve the detection accuracy of lane lines in various scenes, and the improvement effect is most obvious in the case of bad lighting and lane lines curving.Among them, when the lane line is curved, the mIOU with steering angle is 4.4\% higher than that without steering angle, and in the case of bad lighting, the mIOU with steering angle is 4.7\% higher than that without steering angle.

\begin{table}[ht]
\caption{Experimental comparison of lane line detection based on image and steering angle fusion.}
\def\arraystretch{1.3}
\setlength{\tabcolsep}{3.5mm}{
\begin{tabular}{c|c|lll}
\hline
Group         & Steering angle & \multicolumn{1}{c}{mIOU} & \multicolumn{1}{c}{Acc} & \multicolumn{1}{c}{F1} \\ \hline
All           & yes            & \textbf{79.4\%}            & \textbf{99.3\%}           & \textbf{93.5\%}          \\
All           & no             & 76.3\%                     & 97.4\%                    & 89.9\%                   \\ \hline
Good light    & yes            & \textbf{78.4\%}            & \textbf{99.1\%}           & \textbf{89.4\%}          \\
Good light    & no             & 76.1\%                     & 98.4\%                    & 87.8\%                   \\
Bad light     & yes            & \textbf{77.2\%}            & \textbf{99.0\%}           & \textbf{92.1\%}          \\
Bad light     & no             & 72.5\%                     & 96.4\%                    & 88.3\%                   \\ \hline
Curve lane    & yes            & \textbf{76.5\%}            & \textbf{99.2\%}           & \textbf{93.0\%}          \\
Curve lane    & no             & 72.1\%                     & 96.4\%                    & 89.1\%                   \\
Straight lane & yes            & \textbf{77.7\%}            & \textbf{99.4\%}           & \textbf{92.4\%}          \\Straight lane & no             & 76.9\%                     & 99.3\%                    & 91.8\%                   \\ \hline
\end{tabular}}
\label{angle}
\end{table}

In addition, we have reproduced the open-source lane line algorithm in recent years on our dataset, and carried out a comparative experiment. Table \ref{angle2} shows that when we integrate the steering angle into our baseline, all indicators exceed other lane line detection algorithms.Among them, mIOU is 2.1\% higher than CondLane, the current best lane detection algorithm, and 4.3\% higher than SCNN.

\begin{table}[ht]
\caption{Compare the latest open-source lane line detection algorithm.}
\def\arraystretch{1.3}
\setlength{\tabcolsep}{2.7mm}{
\begin{tabular}{c|c|ccc}
\hline
Method              & Backbone  & mIOU & Acc  & F1   \\ \hline
SCNN \cite{pan2018spatial}                & VGG16     & 75.1\% & 96.2\% & 89.1\% \\
RESA \cite{zheng2021resa}               & ResNet50  & 75.3\% & 97.0\% & 90.1\% \\
UFLD \cite{qin2020ultra}                & ResNet18  & 75.2\% & 96.5\% & 89.3\% \\
LaneATT \cite{tabelini2021keep}             & ResNet18  & 76.2\% & 97.1\% & 90.4\% \\
LaneATT \cite{tabelini2021keep}             & ResNet34  & 76.5\% & 97.6\% & 91.0\% \\
LaneATT \cite{tabelini2021keep}             & ResNet122 & 77.1\% & 98.2\% & 91.2\% \\
CondLane \cite{liu2021condlanenet}            & ResNet18  & 76.6\% & 97.2\% & 89.7\% \\
CondLane \cite{liu2021condlanenet}            & ResNet34  & 76.9\% & 97.5\% & 90.7\% \\
CondLane \cite{liu2021condlanenet}            & ResNet101 & 77.3\% & 98.2\% & 91.1\% \\ \hline
LaneNet+Angle (ours) & VGG16         & \textbf{79.4\%} & \textbf{99.3\%} & \textbf{93.5\%} \\ \hline
\end{tabular}}
\label{angle2}
\end{table}

\section{Conclusions}
\label{7}
Most of the existing multi-modal fusion methods use image and point cloud data, which can not avoid the impact of the external environment and the limitations of the sensor itself. However, the vehicle bottom information(steering angle, speed etc) will not have the above factors. It provides real-time and accurate information and provides an intuitive representation of the vehicle's status. In this paper, we first elaborate and analyze some studies of application steering angle and vehicle speed. Secondly, the multi-modal fusion technology is described in detail. Then, we  recommend some public datasets that contain both images and informations of the vehicle. Finally, We describe the current challenges and future research prospects. Our 
survey can provide meaningful references for researchers interested in vehicle information and multi-modal fusion related fields.


\bibliographystyle{IEEEtran}
\bibliography{IEEEabrv,mybib}

\end{document}